\documentclass[11pt]{article}

\usepackage[final]{acl}

\usepackage{times}
\usepackage{latexsym}

\usepackage[T1]{fontenc}

\usepackage[utf8]{inputenc}

\usepackage{microtype}

\usepackage{inconsolata}
\usepackage{stfloats}

\usepackage{graphicx}
\usepackage{booktabs}
\usepackage{array}
\usepackage[table]{xcolor}
\definecolor{oursrow}{HTML}{EAF4FF}
\newcolumntype{C}[1]{>{\centering\arraybackslash}p{#1}}
\usepackage{algorithm}
\usepackage{algpseudocode}
\usepackage{amsmath}
\usepackage[most]{tcolorbox}

\title{Entropy-Regularized Rank-Masked Policy Optimization for\\Test-Time Reinforcement Learning in Code Generation}

\author{Jiacheng Xu, Feng Chen, Xiuneng Xu, Bo An \\
  Nanyang Technological University, Singapore \\
  \texttt{jiacheng005@e.ntu.edu.sg}, \texttt{boan@ntu.edu.sg}}

\begin{document}
\maketitle
\begin{abstract}
Existing methods for test-time reinforcement learning (TTRL) derive rewards from answer-level self-voting on unlabeled test-time tasks with canonical answers, but this breaks down for code generation because programs cannot be compared by surface form and therefore do not directly provide a usable training signal.
To make TTRL applicable to code generation, we propose probe-driven TTRL, which constructs output-free probe inputs from the problem statement, executes candidate programs on these probes, and defines a Probe Consensus Reward (PCR) from the resulting behavioral agreement.
PCR provides a behavioral training signal for open-vocabulary programs, but it is not a fully reliable verifier and remains susceptible to reward hacking through spurious consensus.
We therefore introduce Entropy-Regularized Rank-Masked Policy Optimization (ERPO), which converts low PCR into conservative negative updates through rank masking and controls policy drift with an entropy ceiling.
On coding benchmarks, ERPO substantially improves pass@1 and pass@k in both in-domain adaptation and zero-shot transfer.
\end{abstract}

\begin{figure*}[!b]
  \centering
  \includegraphics[width=\textwidth]{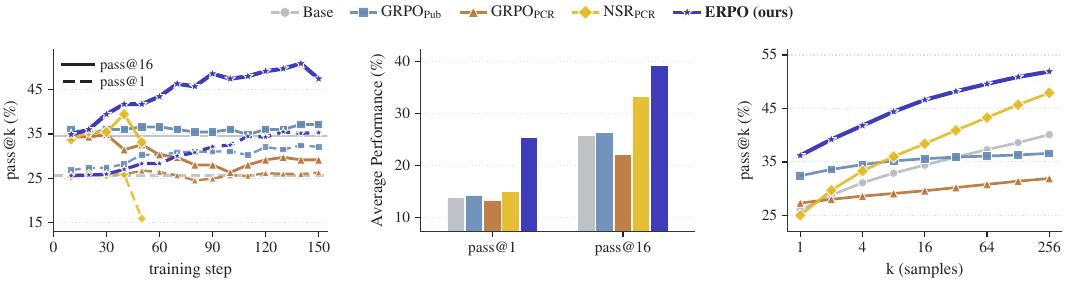}
  \caption{Performance overview. Left: Qwen3-8B per-step LCB results. Middle: Qwen3-4B mean transfer performance across three target benchmarks. Right: Qwen3-4B LCB inference-time scaling.}
  \label{fig:teaser}
\end{figure*}

\section{Introduction}
\label{sec:intro}

As large language models are increasingly trusted with critical tasks, deployment-time response quality has become a central concern~\citep{aghazadeh2026hierarchical}.
A common way to improve quality at test time is inference-time scaling, which spends additional computation without changing model parameters, typically by sampling multiple solutions and selecting one through voting, verification, or search \citep{wang2023selfconsistency,shi2022natural,li2025sstar}.
Such search can be effective, but it is expensive and increases response latency when high-quality answers require many samples.
Test-time reinforcement learning (TTRL) offers a complementary direction: it converts test-time signals into rewards and adapts the model on unlabeled test-time tasks, aiming to improve single-pass response quality at deployment time \citep{zuo2025ttrl,liao2026t3rl,evolrl2025}.
This setting is most natural for tasks such as mathematical reasoning, where sampled solutions can be reduced to directly comparable canonical answers and the majority answer can serve as a pseudo-label \citep{wang2023selfconsistency,zuo2025ttrl}.

For code generation, however, model outputs are open-vocabulary programs rather than canonical final answers, so exact-match voting cannot be directly used to construct a reward \citep{li2022alphacode,shi2022natural}.
Prior training-free inference-time search methods avoid exact matching by comparing execution behavior rather than program text, using execution traces or functional consensus to compare candidate programs \citep{shi2022natural,chen2023codet,launer2026fmv}.
Turning this behavioral comparison into a TTRL reward is nontrivial because test-time adaptation lacks the inputs needed to evaluate generated programs.
The model must therefore construct inputs itself and turn the resulting outputs into a reward signal suitable for reinforcement learning.
A further challenge is that existing TTRL objectives often emphasize pass@1, which can improve single-pass performance and reduce the need for costly training-free search, but can also weaken the pass@k frontier when the gain comes from reduced solution diversity~\citep{evolrl2025}.
This trade-off is especially undesirable in quality-sensitive applications, where pass@k serves as a practical upper bound on what training-free inference-time scaling can extract from the model~\citep{chen2026does,snell2025testtimecompute}.

To make TTRL applicable to code generation while improving both single-pass accuracy and the pass@k frontier, we propose \emph{probe-driven test-time reinforcement learning}.
For each coding problem, the model constructs a set of problem-specific, output-free probe inputs from the problem statement.
It then executes sampled programs on these inputs and defines a Probe Consensus Reward (PCR) from the induced output distributions.
PCR turns open-vocabulary programs into comparable training samples without relying on oracle outputs.
However, because PCR is built from generated inputs and consensus over candidate outputs, it is not a fully reliable verifier.
High consensus may reflect a correct solution, but it may also arise from a common bug or from probes that miss the decisive corner case~\citep{tambon2025bugs,he2025hardtests}.
Directly optimizing PCR can therefore trigger reward hacking and degrade the pass@k frontier while improving pass@1.

Motivated by this observation, we propose \textbf{E}ntropy-Regularized \textbf{R}ank-Masked \textbf{P}olicy \textbf{O}ptimization (\textbf{ERPO}).
Our analysis shows that the PCR signal is asymmetric: low-consensus programs are easier to identify as poor candidates than high-consensus programs are to certify as correct.
Rather than treating high-PCR programs as positive labels, ERPO uses PCR mainly as a negative signal.
Within each candidate group, it masks out the high-ranked half.
The remaining lower-ranked samples typically have negative group-normalized advantages, so the update reduces the likelihood of low-consensus programs while avoiding direct reinforcement of potentially spurious majorities.
A fixed entropy-ceiling regularizer further controls upward entropy drift and preserves stable sampling behavior for pass@k.

We summarize our contributions as follows:

\begin{enumerate}
    \item We formulate probe-driven TTRL for code generation, where output-free inputs generated from the problem statement provide a behavioral interface for rewards without oracle tests or oracle outputs.
    \item We analyze the noise pattern of probe-based PCR rewards and show why direct optimization can be brittle: behavioral consensus is more reliable as a negative signal than as a correctness certificate.
    \item We propose ERPO, a policy optimization algorithm that combines negative rank-masked PCR updates with fixed entropy-ceiling control, improving both pass@1 and pass@k with gains that transfer across coding benchmarks.
\end{enumerate}

\begin{figure*}[t!]
    \centering
    \includegraphics[width=\textwidth]{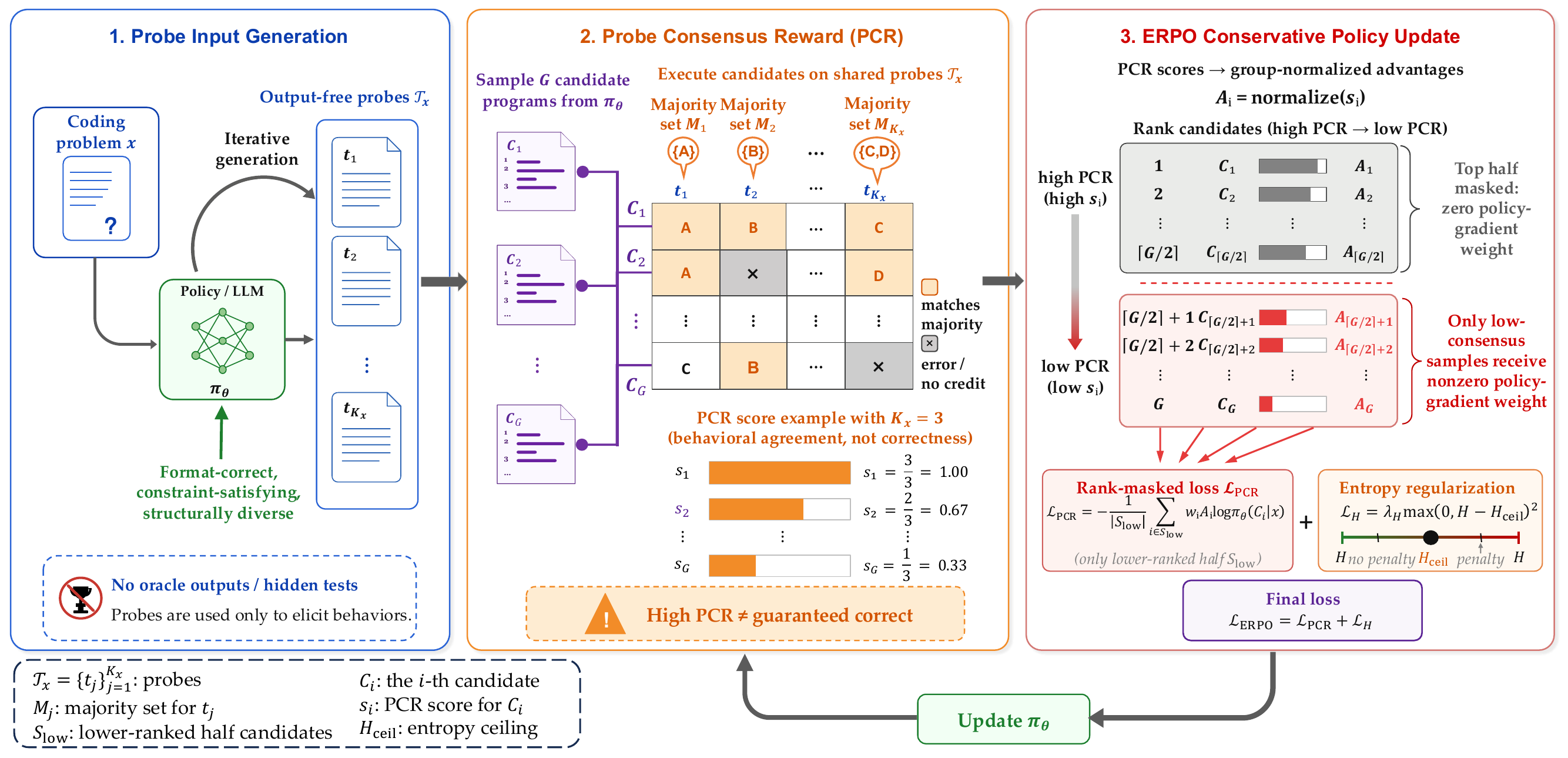}
    \caption{\textbf{Overview of ERPO.} Our method has three stages: (1) generating output-free probe inputs for each problem; (2) executing sampled candidates on the shared probes and scoring them by behavioral agreement (PCR); (3) a rank-masked, entropy-regularized update that uses PCR primarily to suppress low-consensus candidates.}
    \label{fig:method}
\end{figure*}

\section{Preliminaries}
\label{sec:preliminaries}

\subsection{Test-Time Reinforcement Learning}

Reinforcement learning with verifiable rewards (RLVR)~\citep{guo2025deepseek,yu2025dapo} optimizes a policy using scalar rewards from an external verifier.
Given an input $x$ from a training distribution $\mathcal{D}$ and a sampled output $y \sim \pi_{\theta_{\mathrm{old}}}(\cdot \mid x)$, the verifier produces $r(x,y)$, which is converted into an advantage estimate $\hat{A}(x,y)$.
Abstracting away clipping, normalization, and regularization, a generic policy-gradient objective is
\[ 
\mathcal{J}_{\mathrm{RLVR}}(\theta)
=
\mathbf{E}
\left[
\hat{A}(x,y)\log\pi_\theta(y\mid x)
\right].
\]

Test-time reinforcement learning (TTRL) applies this idea during inference, where a model adapts to unlabeled test-time tasks rather than a fixed offline training set.
Because a trusted verifier is unavailable, TTRL must construct a surrogate reward $\tilde{r}(x,y)$.
For canonical problems such as math, this reward is often derived from answer-level self-voting: given $G$ sampled solutions $\{y_i\}_{i=1}^{G}$ with extracted answers $a_i=\mathrm{Ans}(y_i)$, define
\[
a^\star = \mathrm{mode}\{a_i\}_{i=1}^{G},
\]
\[
\tilde{r}_{\mathrm{vote}}(x,y_i)
=
\left\{
\begin{array}{ll}
1, & a_i = a^\star,\\
0, & \mathrm{otherwise}.
\end{array}
\right.
\]
This construction assumes that outputs expose comparable final answers.
For open-vocabulary outputs such as code, semantically equivalent programs may have unrelated surface forms, so reward construction must compare program behavior rather than answer identity.

\subsection{RLVR for Code Generation}

In code generation, the model receives a natural-language problem description $P$ and produces a program $C$ that should satisfy the problem specification.
A standard verifier evaluates $C$ against a test suite $\mathcal{T}=\{(I_j,O_j)\}_{j=1}^{m}$, where $I_j$ is a test input and $O_j$ is the corresponding oracle output.
When an ideal test suite $\mathcal{T}^{\star}$ is available, the code reward can be written as \citep{yu2025dapo}
\[
R_{\mathrm{code}}(C,\mathcal{T}^{\star})
=
\left\{
\begin{array}{ll}
1, & \begin{array}{l}
\forall (I_j,O_j) \in \mathcal{T}^{\star},\\
\mathrm{Exec}(C,I_j)=O_j,
\end{array}\\
0, & \mathrm{otherwise}.
\end{array}
\right.
\]
Here $\mathrm{Exec}(C,I_j)$ denotes the output produced by executing program $C$ on input $I_j$.
This formulation makes test quality central to training and evaluation: the reward is reliable only when the test suite is complete and the oracle outputs are trusted.

When TTRL is applied to code generation, however, such a trusted test suite is not available for the target problem.
The problem does not provide test inputs and corresponding oracle outputs from which to obtain a reliable reward signal.
As a result, applying the standard code reward is impossible.
The central challenge is to build a surrogate reward that can compare candidate programs without private test cases or ground-truth outputs, while providing a useful signal for test-time policy updates.\looseness=-1

\section{Method}
\label{sec:method}

\paragraph{Overview.}
We study test-time reinforcement learning for code generation, where a model adapts to target coding problems without reliable input-output tests for full program verification.
Beyond missing labels, code generation lacks canonical answers for self-voting: generated programs are open-vocabulary outputs that cannot be compared by surface form for majority voting.
The model must therefore construct training signals from program behavior, instead.

Our method addresses this difficulty in three stages.
For each problem, we first generate a set of distinct, output-free probe inputs from the problem statement, as described in Section~\ref{sec:input_generation}.
Because corresponding oracle outputs are unavailable, these probes expose behavior rather than verify correctness.
During training or evaluation, we sample multiple candidate programs, execute them on the shared probes, and compute a Probe Consensus Reward (PCR) from output consensus (Section~\ref{sec:pcr}).
Finally, we introduce Entropy-Regularized Rank-Masked Policy Optimization (ERPO).
Instead of directly maximizing this noisy reward, ERPO applies a rank-masked negative update to the lower-PCR part of each candidate group, while assigning high-PCR responses zero policy-gradient weight. The full objective is given in Section~\ref{sec:optimization}.

\subsection{Probe Input Generation}
\label{sec:input_generation}

Let $\mathcal{D}=\{x_i\}_{i=1}^{m}$ be the target set of coding problems.
For each problem $x$, the model observes the natural-language problem statement without access to any test suite for reliable evaluation.
Because generated programs are open-vocabulary text outputs and do not support consistency voting by surface form, we compare programs by execution behavior on shared inputs.

Unlike supervised code RL with input-output test suites for verifying code correctness, these shared inputs must be constructed at adaptation time and have no oracle outputs.
Before optimization, for each problem $x$, we ask the model to generate $K_x$ raw test inputs and form the probe set $\mathcal{T}_x=\{t_j\}_{j=1}^{K_x}$.
This set is generated once and reused across training steps.
Generation proceeds iteratively: at iteration $j$, the prompt contains the problem statement and all previously generated inputs $\{t_1,\ldots,t_{j-1}\}$.
The model produces one new input $t_j$ that is format-correct, constraint-satisfying, and structurally different from the previous ones.
The prompt template is provided in Appendix~\ref{app:probe-input-prompt}.
Thus $\mathcal{T}_x$ encourages distinct execution regimes and provides label-free common execution points for behavioral comparison, not tests with trusted outputs.
The reward below depends only on candidate outputs induced by $\mathcal{T}_x$, without any reference outputs.\looseness=-1

\subsection{Probe Consensus Reward}
\label{sec:pcr}

Execution-based consensus has been used for code selection and functional voting.
Here, we adopt this idea as a test-time training reward over generated, output-free probes.
Given a problem $x$, we sample a group of $G$ candidate programs $\{C_i\}_{i=1}^{G}$ from the current policy.
For each probe input $t_j \in \mathcal{T}_x$, we execute each candidate $C_i$ and record its result $z_{ij}$.
If execution fails, the candidate receives no credit for that input and is excluded from that input's majority computation.

For each probe input $t_j$, we count only valid outputs.
The count
\[
n_j(o)=
\sum_{i=1}^{G}\mathbf{1}\{z_{ij}=o\}
\]
is the number of candidates that produce output $o$ on $t_j$.
The tied majority set is
\[
\mathcal{M}_j =
\{o: n_j(o)=\max_{o'} n_j(o')\}.
\]
Ties are credited to all tied outputs.

Candidate $C_i$ receives the PCR score
\[
s_i =
\frac{1}{K_x}
\sum_{j=1}^{K_x}
\mathbf{1}\{z_{ij}\in\mathcal{M}_j\},
\]
which is the fraction of probe inputs for which its output belongs to the majority set.
Thus $s_i\in[0,1]$ measures behavioral agreement with the candidate group on the generated probes, without oracle outputs or any claim that the majority output is correct.

PCR is useful but not a correctness verifier.
A high score can still correspond to an incorrect program when many candidates share a bug or the probes miss the relevant corner case.
Conversely, a correct program can receive a low score if it follows a rare but valid behavior while most sampled programs agree on a wrong one.
Our policy optimization algorithm therefore uses PCR conservatively.\looseness=-1

\subsection{Entropy-Regularized Rank-Masked Policy Optimization Algorithm}
\label{sec:optimization}

A direct GRPO update on PCR scores would reinforce high-scoring candidates and penalize low-scoring ones.
This is unsafe because high PCR is not a reliable correctness certificate.
In practice, direct optimization can initially improve pass@1 by reducing sampling diversity, but continued training can overfit to the reward and degrade the upper-bound performance captured by pass@k.
ERPO therefore does not treat high-PCR candidates as targets.
Instead, it uses PCR through a rank mask and fixed entropy-ceiling regularization.

For a group of $G$ candidates from the same problem, let $s_i$ be the PCR score defined above.
We first compute the standard group-normalized GRPO advantage \citep{shao2024deepseekmath}:
\[
A_i =
\frac{s_i-\mu_x}{\sigma_x},
\quad
\mu_x=\frac{1}{G}\sum_{i=1}^{G}s_i,
\]
where $\sigma_x$ is the within-group standard deviation of the PCR scores.
This centers the signal within each problem, so the update depends on relative behavioral support rather than the absolute PCR scale.\looseness=-1

ERPO then masks the GRPO advantage by rank.
Within each group, candidates are ranked by $A_i$ in descending order.
The top $\lceil G/2\rceil$ candidates are masked out, and the loss keeps only the signed advantages of the lower-ranked half:
\[
\widetilde{A}_i =
A_i \cdot
\mathbf{1}\{
\mathrm{rank}^{\downarrow}_x(i)>\lceil G/2\rceil
\}.
\]
Here $\mathrm{rank}^{\downarrow}_x(i)=1$ denotes the candidate with the largest advantage within problem $x$.
This mask turns PCR into a conservative negative signal.
High-PCR candidates are excluded from positive policy-gradient updates, while the retained lower-ranked candidates usually have $A_i<0$ and thus reduce the likelihood of low-consensus programs.

The policy component of ERPO is the rank-masked PCR loss $\mathcal{L}_{\mathrm{PCR}}$.
Ignoring clipping for notation, its gradient direction is proportional to
\[
-\sum_{i=1}^{G}
\widetilde{A}_i
\sum_{t=1}^{|C_i|}
\nabla_{\theta}
\log \pi_{\theta}(C_{i,t}\mid x,C_{i,<t}).
\]
Here $C_{i,<t}$ denotes the prefix of candidate program $C_i$ before token $t$.

Rank-masked updates can still destabilize training when they drive policy entropy upward without bound.
We therefore add a fixed entropy-ceiling regularizer.
Let $H_\theta$ be the token-level policy entropy averaged over the current actor update, and let $H_{\mathrm{ceil}}$ be a fixed entropy ceiling chosen before adaptation.
We penalize only violations of this ceiling:\looseness=-1
\[
\mathcal{L}_{H}
=
\lambda_{H}
\max\left(0, H_\theta-H_{\mathrm{ceil}}\right)^2.
\]
Because the regularizer is inactive below $H_{\mathrm{ceil}}$ and grows quadratically above it, it leaves lower-entropy updates unpenalized and suppresses excessive entropy growth.
Putting the two components together, ERPO minimizes the following objective, with the update summarized in Algorithm~\ref{alg:erpo}:
\[
\mathcal{L}_{\mathrm{ERPO}}
=
\mathcal{L}_{\mathrm{PCR}}
+ \mathcal{L}_{H}.
\]

\begin{algorithm}[t]
\caption{Entropy-Regularized Rank-Masked Policy Optimization (ERPO)}
\label{alg:erpo}
\begin{algorithmic}[1]
\State \textbf{Input:} problem set $\mathcal{D}$, precomputed probe sets $\{\mathcal{T}_x\}_{x\in\mathcal{D}}$, policy $\pi_\theta$, group size $G$, entropy-ceiling parameters $(\lambda_H,H_{\mathrm{ceil}})$
\For{each training step}
    \State Sample problem $x \sim \mathcal{D}$
    \State Retrieve the precomputed probe set $\mathcal{T}_x$
    \State Set $\pi_{\theta_{\mathrm{old}}}\leftarrow\pi_\theta$
    \State Sample programs $C_i\sim\pi_{\theta_{\mathrm{old}}}(\cdot\mid x)$ for $i=1,\ldots,G$
    \State Execute $\{C_i\}_{i=1}^{G}$ on the probe set $\mathcal{T}_x$
    \State Compute PCR scores $s_i$ for $i=1,\ldots,G$
    \State Compute group advantages $\{A_i\}_{i=1}^{G}$
    \State Apply the rank mask to obtain $\{\widetilde{A}_i\}_{i=1}^{G}$
    \State Compute $\mathcal{L}_{\mathrm{ERPO}}=\mathcal{L}_{\mathrm{PCR}}+\mathcal{L}_{H}$
    \State Update $\theta$ by one gradient step on $\mathcal{L}_{\mathrm{ERPO}}$
\EndFor
\State \textbf{Output:} updated policy $\pi_\theta$
\end{algorithmic}
\end{algorithm}

\section{Experiments}
\label{sec:experiments}

\subsection{Experimental Setup}
We use Qwen3-4B and Qwen3-8B as base models \citep{yang2025qwen3}.
The main experiments adapt on LiveCodeBench (LCB) \citep{jain2025livecodebench} and evaluate zero-shot transfer to CodeContests (CC) \citep{li2022alphacode}, CodeForces (CF) \citep{cure2025}, and TACO \citep{li2023taco}.
As unlabeled-reward baselines, GRPO$_{\mathrm{PCR}}$ and NSR$_{\mathrm{PCR}}$ access only problem statements, without provided test cases, and update the model with GRPO or NSR using PCR rewards \citep{shao2024deepseekmath,zhu2025negative}.
In particular, NSR$_{\mathrm{PCR}}$ uses PCR as a negative-only signal by penalizing low-score samples.
ERPO follows the same setting, using the objective in Section~\ref{sec:optimization}.
We also include GRPO$_{\mathrm{Pub}}$, a more informative baseline that observes the complete inputs and outputs of public test cases at test time for reward computation.
This verifier is reliable but incomplete: failing a public test implies incorrectness, but passing all public tests may still fail hidden tests.
Implementation details for all methods are provided in Appendix~\ref{app:experimental-configuration}.

\paragraph{Evaluation protocol.}
We report two settings.
In-domain adaptation updates only on unlabeled problem statements from the target benchmark: during training, the model never observes any evaluation test input and its oracle output. The complete evaluation suite is used only post hoc to compute pass@k (P@k) \citep{chen2021evaluating}.
The transfer experiment evaluates reusability: we adapt once on unlabeled LCB problems, then test the checkpoint on CC, CF, and TACO without target-benchmark updates. 
This setting distinguishes benchmark-specific optimization from broader capability gains across the code-generation domain.
Unless otherwise noted, we use the final checkpoint from the prespecified schedule, without access to hidden-test pass@k.

\subsection{Main Results}
\label{sec:train}

\begin{table*}[t]
  \centering
  \small
  \setlength{\tabcolsep}{4.0pt}
  \renewcommand{\arraystretch}{1.12}
  \caption{Main code-generation results. Mean averages the three transfer targets. Values are percentages.}
  \label{tab:main_results}
  \begin{tabular}{@{}llcc@{\hspace{10pt}}*{8}{c}@{}}
    \toprule
    & & \multicolumn{2}{c}{In-Domain Adapt.} & \multicolumn{8}{c}{Zero-Shot Transfer} \\
    \cmidrule(lr){3-4}\cmidrule(l){5-12}
    Model & Method & \multicolumn{2}{c}{LCB} & \multicolumn{2}{c}{CC} & \multicolumn{2}{c}{CF} & \multicolumn{2}{c}{TACO} & \multicolumn{2}{c}{Mean} \\
    \cmidrule(lr){3-4}\cmidrule(lr){5-6}\cmidrule(lr){7-8}\cmidrule(lr){9-10}\cmidrule(l){11-12}
    & & P@1 & P@16 & P@1 & P@16 & P@1 & P@16 & P@1 & P@16 & P@1 & P@16 \\
    \midrule
    Qwen3-4B & Base & 26.0 & 34.4 & 25.1 & 42.3 & 9.8 & 21.8 & 6.4 & 13.1 & 13.8 & 25.7 \\
    & GRPO$_{\mathrm{Pub}}$ & \underline{33.8} & 35.4 & 26.2 & 41.0 & 10.7 & 24.4 & 6.1 & 13.6 & 14.3 & 26.3 \\
    & GRPO$_{\mathrm{PCR}}$ & 27.8 & 30.9 & 26.0 & 36.4 & 8.5 & 20.1 & 5.5 & 9.8 & 13.3 & 22.1 \\
    & NSR$_{\mathrm{PCR}}$ & 25.7 & \underline{39.4} & \underline{27.4} & \underline{51.0} & \underline{10.8} & \underline{32.1} & \underline{6.6} & \underline{16.6} & \underline{14.9} & \underline{33.2} \\
    \rowcolor{oursrow}
    & \textbf{ERPO} & \textbf{36.7} & \textbf{46.3} & \textbf{42.1} & \textbf{58.2} & \textbf{20.7} & \textbf{38.5} & \textbf{13.3} & \textbf{21.3} & \textbf{25.4} & \textbf{39.3} \\
    \midrule
    Qwen3-8B & Base & 25.6 & 34.6 & 27.2 & 45.6 & 9.3 & 22.5 & 7.2 & 14.6 & 14.6 & 27.6 \\
    & GRPO$_{\mathrm{Pub}}$ & \underline{32.1} & 37.1 & 28.3 & 41.8 & \underline{11.5} & 27.2 & \underline{7.6} & 13.2 & \underline{15.8} & 27.4 \\
    & GRPO$_{\mathrm{PCR}}$ & 26.3 & 29.1 & 27.2 & 41.0 & 10.1 & 22.1 & 7.0 & 13.1 & 14.8 & 25.4 \\
    & NSR$_{\mathrm{PCR}}$ & 25.9 & \underline{39.4} & \underline{28.8} & \underline{52.3} & 10.3 & \underline{30.0} & 7.1 & \underline{17.3} & 15.4 & \underline{33.2} \\
    \rowcolor{oursrow}
    & \textbf{ERPO} & \textbf{35.1} & \textbf{50.9} & \textbf{39.4} & \textbf{57.7} & \textbf{17.5} & \textbf{37.3} & \textbf{11.8} & \textbf{21.3} & \textbf{22.9} & \textbf{38.8} \\
    \bottomrule
  \end{tabular}
\end{table*}

Table~\ref{tab:main_results} shows ERPO's clear advantage. In-domain on LCB, ERPO improves both pass@1 and pass@16, raising single-pass accuracy and the pass@k frontier, i.e., the coverage of correct solutions under repeated sampling. Its gains also transfer to CC, CF, and TACO, showing cross-benchmark generalization. GRPO$_{\mathrm{PCR}}$, which directly optimizes PCR, performs worst: on LCB it obtains only a small pass@1 gain with a large pass@16 drop, and the pass@1 gain disappears on transfer benchmarks, suggesting overfitting to noisy PCR patterns. GRPO$_{\mathrm{Pub}}$, with ground-truth public tests, is second-best for pass@1, but its pass@16 gains are marginal and its 8B transfer gains barely persist, indicating reward hacking on the adaptation source. NSR$_{\mathrm{PCR}}$ improves pass@16 but shows little pass@1 gain because repeated penalties on negative samples sharply increase entropy, so its pass@16 gain mainly comes from diversity. As shown in the appendix, NSR$_{\mathrm{PCR}}$ soon suffers entropy explosion and policy collapse. ERPO is therefore the only adaptation method that substantially improves both pass@1 and pass@16, using a small amount of fully unlabeled test-time data to exploit the PCR signal while avoiding its noisy patterns.

\subsection{Analysis of the PCR Score}
\label{sec:pcr_analysis}

We analyze PCR to clarify when the surrogate reward is reliable and how its noise affects policy optimization.
Table~\ref{tab:pcr_score_analysis} compares PCR with hidden-test correctness on LiveCodeBench.
Hidden tests are used only for analysis and are unavailable during test-time adaptation.
PCR has useful ranking power, with AUC 0.790 for Qwen3-4B and 0.750 for Qwen3-8B, but its calibration is highly asymmetric.
Low PCR is a strong negative signal: candidates with $s \le 0.5$ fail hidden tests more than 91\% of the time for both model sizes.
High PCR, however, is not a correctness certificate: even candidates with $s = 1.0$ are wrong 44.2\% of the time for Qwen3-4B and 51.9\% for Qwen3-8B.

\begin{table}[t]
  \centering
  \small
  \setlength{\tabcolsep}{4.8pt}
  \renewcommand{\arraystretch}{1.12}
  \caption{PCR score $s$ calibration on LiveCodeBench. $y_{\mathrm{hid}}$ denotes hidden-test correctness. Fail rate is the fraction of candidates that do not pass hidden tests.}
  \label{tab:pcr_score_analysis}
  \begin{tabular}{@{}lcc@{}}
    \toprule
    Metric & Qwen3-4B & Qwen3-8B \\
    \midrule
    AUC$(s_{\mathrm{PCR}}, y_{\mathrm{hid}})$ & 0.790 & 0.750 \\
    Fail rate among $s \le 0.5$ & 92.0\% & 91.3\% \\
    Fail rate among $s \le 0.25$ & 91.1\% & 91.0\% \\
    Fail rate among $s \ge 0.9$ & 50.7\% & 54.6\% \\
    Fail rate among $s = 1.0$ & 44.2\% & 51.9\% \\
    \bottomrule
  \end{tabular}
\end{table}

This asymmetry explains why direct PCR optimization is brittle.
A GRPO update can turn relatively high-PCR candidates into positive targets, reinforcing consensual wrong programs when a group contains no hidden-correct solution.
ERPO uses PCR mainly as a conservative negative signal, penalizing low-consensus programs while avoiding positive updates on high-consensus false positives.

\subsection{Inference-Time Scaling}
\label{sec:inference}

\begin{table*}[t]
  \centering
  \small
  \setlength{\tabcolsep}{3.0pt}
  \renewcommand{\arraystretch}{1.12}
  \caption{Inference-time pass@k on Qwen3-4B. All values are percentages.}
  \label{tab:inference_time_scaling}
  \begin{tabular}{@{}l*{9}{C{0.072\textwidth}}@{}}
    \toprule
    Method & P@1 & P@2 & P@4 & P@8 & P@16 & P@32 & P@64 & P@128 & P@256 \\
    \midrule
    \multicolumn{10}{l}{\textit{LiveCodeBench}} \\
    Base & 26.0 & 28.8 & 31.1 & 32.9 & 34.4 & 35.9 & 37.3 & 38.6 & 40.1 \\
    GRPO$_{\mathrm{Pub}}$ & \underline{32.4} & \underline{33.6} & \underline{34.5} & 35.2 & 35.6 & 35.9 & 36.1 & 36.3 & 36.6 \\
    GRPO$_{\mathrm{PCR}}$ & 27.3 & 28.0 & 28.6 & 29.1 & 29.6 & 30.2 & 30.8 & 31.4 & 31.9 \\
    NSR$_{\mathrm{PCR}}$ & 25.0 & 29.7 & 33.3 & \underline{36.0} & \underline{38.4} & \underline{40.9} & \underline{43.3} & \underline{45.7} & \underline{47.9} \\
    \rowcolor{oursrow}
    ERPO & \textbf{36.2} & \textbf{39.2} & \textbf{41.8} & \textbf{44.4} & \textbf{46.6} & \textbf{48.2} & \textbf{49.6} & \textbf{50.9} & \textbf{51.9} \\
    \midrule
    \multicolumn{10}{l}{\textit{CodeContests}} \\
    Base & 25.0 & 29.3 & 33.5 & 37.5 & 41.2 & 44.6 & 47.3 & 49.2 & 50.3 \\
    GRPO$_{\mathrm{Pub}}$ & \textbf{42.2} & \underline{43.3} & \underline{44.2} & 45.0 & 45.6 & 46.1 & 46.6 & 47.1 & 47.7 \\
    GRPO$_{\mathrm{PCR}}$ & 27.1 & 28.6 & 30.4 & 32.5 & 34.7 & 36.7 & 38.4 & 39.8 & 40.7 \\
    NSR$_{\mathrm{PCR}}$ & 27.3 & 34.8 & 41.1 & \underline{46.3} & \underline{50.9} & \underline{54.9} & \underline{58.3} & \underline{61.4} & \underline{64.1} \\
    \rowcolor{oursrow}
    ERPO & \underline{41.5} & \textbf{46.5} & \textbf{50.5} & \textbf{53.8} & \textbf{56.9} & \textbf{59.8} & \textbf{62.4} & \textbf{64.5} & \textbf{65.9} \\
    \bottomrule
  \end{tabular}
\end{table*}

\begin{table}[t]
  \centering
  \small
  \setlength{\tabcolsep}{8.0pt}
  \renewcommand{\arraystretch}{1.12}
  \caption{Best-of-$n$ accuracy of training-free search for Qwen3-4B on LiveCodeBench.}
  \label{tab:bon_vs_oracle}
  \begin{tabular}{@{}lcc@{}}
    \toprule
    Method & CodeT-BoN & RM-BoN \\
    \midrule
    Base, $n{=}1$  & 26.0 & 26.0 \\
    Base, $n{=}4$  & 28.6 & 29.7 \\
    Base, $n{=}8$  & 28.0 & 28.0 \\
    Base, $n{=}16$ & 28.0 & 28.0 \\
    Base, $n{=}32$ & 29.1 & 27.4 \\
    \midrule
    \rowcolor{oursrow}
    \textbf{ERPO} & \multicolumn{2}{c}{\textbf{36.2}\ \ (single sample, no search cost)} \\
    \bottomrule
  \end{tabular}
  \vspace{-0.5em}
\end{table}

For test-time training, pass@k is crucial \citep{chen2021evaluating} because training-free inference-time search needs many samples to obtain stronger submissions and is bounded by the underlying pass@k frontier. We evaluate inference-time scaling for checkpoints adapted on each of LiveCodeBench and CodeContests by sampling 256 solutions per problem and reporting pass@k as $k$ varies (Table~\ref{tab:inference_time_scaling}). ERPO substantially improves pass@k on both datasets at every $k$, showing better single-pass accuracy and a stronger frontier for test-time search. In contrast, GRPO$_{\mathrm{Pub}}$ and GRPO$_{\mathrm{PCR}}$ severely sacrifice pass@k headroom. NSR$_{\mathrm{PCR}}$ also improves pass@k, but its gains grow with $k$ and remain limited at pass@1, suggesting that they mainly reflect sampling diversity rather than higher single-pass accuracy.

On LiveCodeBench, ERPO's pass@1 exceeds the training-free Base model's pass@32, and on CodeContests exceeds Base pass@16.
Existing inference-time search methods must select among candidates using a reward model or generated tests, which adds search cost and remains bounded by pass@k, whereas ERPO adapts the model itself and reduces the deployment search budget.
The pass@k values in Table~\ref{tab:inference_time_scaling} are oracle upper bounds, assuming a perfect selector can pick any correct sample out of $k$, whereas deployment must submit one solution using only test-time information.
Table~\ref{tab:bon_vs_oracle} therefore reports Best-of-$n$ accuracy on the Qwen3-4B base model with two representative selectors: CodeT-BoN (\citealp{chen2023codet}), which samples $M{=}32$ LLM-generated $(\text{input},\text{expected output})$ test cases per problem from the base model and ranks $n$ candidate solutions by pass-pattern cluster size, and RM-BoN, which scores the $n$ candidates with Skywork-Reward-Llama-3.1-8B-v0.2~\citep{liu2024skywork}.
Both selectors stagnate below 30\% even at $n=32$, whereas ERPO reaches 36.2\% with one sample and no inference-time search, surpassing every Best-of-$n$ configuration.
This gap shows that current selectors cannot realize much of the oracle pass@k headroom, while test-time adaptation captures it directly.\looseness=-1

\subsection{Hyperparameter Ablation}
\label{sec:hp_ablation}

\begin{figure}[t]
  \centering
  \includegraphics[width=\linewidth]{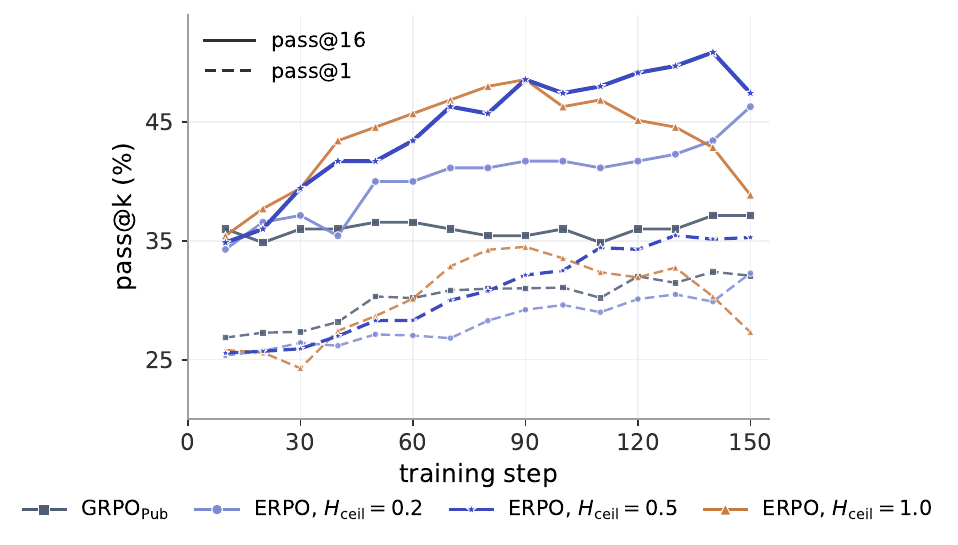}
  \caption{Entropy ceiling ablation on LiveCodeBench.}
  \label{fig:ceiling_ablation}
  \vspace{-0.5em}
\end{figure}

We ablate the ERPO entropy ceiling with $H_{\mathrm{ceil}}\in\{0.2,0.5,1.0\}$ on Qwen3-8B / LiveCodeBench, holding all other hyperparameters fixed as in Appendix~\ref{app:experimental-configuration}.
Figure~\ref{fig:ceiling_ablation} shows per-step pass@1 (dashed) and pass@16 (solid), with GRPO$_{\mathrm{Pub}}$ as a non-PCR reference.
All three values improve pass@1 and pass@16, indicating that the ERPO objective is the main driver of the gains.
The ceiling affects training efficiency and final stability: lower values slow training, whereas an overly high value accelerates early pass@k improvement but degrades quickly after its peak.
Thus, higher entropy does not necessarily imply a stronger pass@k frontier.
We therefore fix $H_{\mathrm{ceil}}{=}0.5$, which best balances efficiency and stability, across all training experiments without model- or dataset-specific tuning.
Ablations on the rank-mask percentile and generated probe count are provided in Appendices~\ref{sec:rank_mask_sensitivity} and~\ref{sec:probe_count_sensitivity}, respectively.\looseness=-1

\section{Related Work}
\label{sec:related}

\subsection{Unsupervised RLVR and Test-Time Reinforcement Learning}

Reinforcement learning with verifiable rewards is most reliable when outputs can be scored by a trusted verifier \citep{yu2025dapo}.
Unsupervised RLVR and test-time reinforcement learning weaken this assumption by deriving rewards from signals available at adaptation time, often through answer-level agreement or self-consistency \citep{zuo2025ttrl,wang2023selfconsistency,liao2026t3rl,he2026urlvr}.
Such rewards are natural for tasks with canonical final answers, but code generation lacks this structure: correct programs can differ substantially in surface form, so reward construction must compare behavior rather than text.

Execution-based comparison offers a natural behavioral interface for code, and functional-voting or label-free code-training methods use candidate outputs on shared inputs to support consensus-based test-time training, supervision from sampled programs and tests, and co-evolution of coders and testers through execution feedback \citep{launer2026fmv,zerocoder2026,cure2025}.
Our work follows this execution-based direction but uses it for probe-driven TTRL: we construct output-free probes from the problem statement, treat the induced agreement as a noisy PCR reward, and use ERPO to avoid directly reinforcing high-consensus samples whose correctness is not certified.

\subsection{Inference-Time Scaling for Code Generation}

Inference-time scaling improves code generation with more test-time compute, typically by sampling many candidate programs and selecting or revising them with an auxiliary execution signal~\citep{snell2025testtimecompute}.
MBR-Exec selects programs at inference by behavioral agreement on test inputs \citep{shi2022natural}.
CodeT and TCS generate additional tests and use execution results for solution selection \citep{chen2023codet,xu2026tcs}.
AlphaCode and later systems scale code generation through large-scale sampling, generated tests, functional decomposition, pairwise execution comparisons, or cluster-based selection \citep{li2022alphacode,chen2024funcoder,li2025sstar,samadi2025gencluster}.

These methods use execution to guide inference-time generation and selection.
Selection alone does not improve the model's generation capability and is bounded by candidate coverage, measured by pass@k under the corresponding generation procedure \citep{chen2021evaluating,chen2023codet}.
Test-time policy optimization can improve this coverage by changing the generation distribution, but may also reinforce incorrect consensus across updates, propagating errors beyond the current problem.
PCR therefore uses execution not as a correctness certificate, but as a behavioral reward interface for test-time adaptation.
ERPO optimizes under this interface with rank masking and fixed entropy-ceiling control, aiming to improve pass@1 while preserving the pass@k frontier and providing a stronger base model for current training-free inference-time scaling algorithms.

\section{Conclusion}
\label{sec:conclusion}

This work revisits test-time reinforcement learning from the perspective of deployment, where improving response quality by spending more computation on inference-time search is costly and can increase latency.
TTRL is attractive because it adapts a model on unlabeled test-time tasks, but code generation exposes two obstacles that standard answer-level voting does not address.
Programs are open-vocabulary outputs, so they do not provide directly comparable canonical answers for reward construction.
At the same time, objectives that primarily improve pass@1 can weaken the pass@k frontier, limiting what quality-sensitive inference-time search can recover from the adapted model.

To address these issues, we propose probe-driven TTRL for code generation.
By constructing output-free probe inputs and defining a Probe Consensus Reward (PCR), our framework makes TTRL for code possible without oracle outputs or directly comparable responses.
We further introduce Entropy-Regularized Rank-Masked Policy Optimization (ERPO), a policy optimization algorithm for this noisy reward, which substantially improves both pass@1 and pass@k.
The adapted model can serve as a stronger test-time base policy: it supports lower-cost, higher-quality single-pass responses in latency-sensitive scenarios, while also providing a better performance frontier for search in quality-sensitive scenarios.
These properties make probe-driven TTRL practically relevant for model deployment and worthy of broader study.

\section*{Limitations}
\label{sec:limitation}

Our empirical evaluation is limited to competitive-programming-style coding benchmarks. The proposed framework, however, is not inherently restricted to this setting. In principle, it can be applied to broader code-generation tasks where candidate code can be executed and assessed through unit tests or other executable checks. We do not evaluate project-level code generation in this work because such settings require substantially more engineering infrastructure for reliable execution, training, and evaluation, along with much larger computational resources. We therefore leave project-level and industrial-scale code-generation scenarios as an important direction for future work.

\bibliography{custom}

\clearpage
\appendix
\section{Experimental Configuration}
\label{app:experimental-configuration}

All our RL experiments adopt the verl~\citep{sheng2024hybridflow} framework with vLLM rollouts and an FSDP actor.
The two backbones are Qwen3-4B and Qwen3-8B in non-thinking mode.
Unless noted otherwise, every method shares the optimization and sampling settings below.
Only the reward source, loss term, and method-specific hyperparameters change between methods.

\paragraph{Optimization.}
We use AdamW with a constant learning rate of $5\!\times\!10^{-7}$ after a $20$-step linear warmup.
The training batch is $16$ prompts; each prompt produces $G=16$ on-policy rollouts, giving a $16{\times}16=256$-rollout step.
PPO mini-batches contain $4$ prompts and we run a single inner PPO epoch per global step. The maximum number of global steps is $150$.
Unless otherwise specified, we evaluate the final checkpoint from the prespecified training schedule, or the last pre-collapse checkpoint for methods that undergo catastrophic policy collapse before the end of training.

\paragraph{Sampling.}
A single sampling profile is shared between RL rollouts and validation rollouts.
For Qwen3 models in non-thinking mode, generation uses $\texttt{top\_p}=0.8$, $\texttt{top\_k}=20$, and $\texttt{min\_p}=0$. For sequence lengths, we use $\texttt{max\_response\_length}=8192$ in experiments.
P@1 and P@16 in the main and transfer tables (Tables~\ref{tab:main_results} and~\ref{tab:cc_transfer_4b}) are computed from $16$ validation samples per problem.

\paragraph{Method-specific hyperparameters.}
For PCR-based methods, the probe sets are precomputed once for each problem and reused across training steps.
The same probe set is shared by GRPO$_{\mathrm{PCR}}$, NSR$_{\mathrm{PCR}}$, and ERPO within each backbone and benchmark setting.
For all experiments, we generate $10$ additional probe inputs per problem. PCR is computed over the deduplicated union of these generated probes and the example inputs in the problem statement.
\begin{itemize}
    \item \textbf{GRPO$_{\mathrm{Pub}}$}: We use standard GRPO with public-binary rewards and a KL coefficient of $10^{-3}$. The reward is $1$ only when the candidate program passes all provided public tests, and $0$ otherwise.
    \item \textbf{GRPO$_{\mathrm{PCR}}$}: We use standard GRPO with pointwise PCR rewards and KL coefficient $10^{-3}$. The PCR score is used directly as the scalar GRPO reward.
    \item \textbf{NSR$_{\mathrm{PCR}}$}: We use pointwise PCR rewards with KL coefficient $10^{-3}$. For a raw PCR score $s_i$, we define the NSR reward $r_i=2(s_i-0.5)\in[-1,1]$ and keep only samples with $r_i<0$ (equivalently, $s_i<0.5$) for the actor update.
    \item \textbf{ERPO}: We use pointwise PCR rewards with KL coefficient $10^{-3}$. ERPO computes the group-normalized GRPO advantage on the PCR scores and masks the top $\lceil G/2\rceil$ candidates in each group. The remaining lower-PCR candidates usually have negative normalized advantages, so the actor update penalizes them rather than reinforcing them, while high-PCR candidates receive zero positive policy-gradient weight. The fixed entropy-ceiling term uses $\lambda_H=0.5$ and $H_{\mathrm{ceil}}=0.5$ with actor entropy computation enabled.
\end{itemize}

\paragraph{Inference-time scaling.}
For the inference-scaling experiment, we resample $N=256$ rollouts per problem from each method's checkpoint with the same sampling profile and report pass@k as a function of $k\in\{1,2,4,8,16,32,64,128,256\}$.

\paragraph{Compute resources.}
All RL training runs were executed on eight NVIDIA A100 GPUs with 80GB memory per GPU.
A complete RL training run for one algorithm takes approximately 12 hours; the inference-time scaling experiment takes about 3 hours to sample 256 times for every 100 problems.

\paragraph{Datasets.}
We use LiveCodeBench v6 for all experiments involving LiveCodeBench.
For all other benchmarks, we use the complete official releases.

\paragraph{Execution environment.}
Candidate programs were executed in isolated subprocesses with per-execution timeout and resource limits.
Network access was disabled, and compilation errors, runtime errors, and timeouts were treated as execution failures.
Failed executions received no credit for the corresponding probe and were excluded from that probe's majority computation.

\paragraph{Artifact licenses and terms.}
We use publicly available research artifacts, including the coding benchmarks, model checkpoints, and software frameworks described in Section~\ref{sec:experiments} and Appendix~\ref{app:experimental-configuration}.
We cite the original creators of these artifacts and use them under their respective licenses and terms of use.
Our experiments use these artifacts for research evaluation and adaptation only, and we do not redistribute the original datasets or model checkpoints beyond what is permitted by their original licenses.

\section{Probe Input Generation Prompt}
\label{app:probe-input-prompt}

Section~\ref{sec:input_generation} constructs the output-free probe set for each problem using an iterative chat prompt.
At iteration $j$, the prompt is instantiated with the problem statement and the $j-1$ probe inputs already generated for the same problem.
The same prompt structure is used for all probe sets reported in our experiments.
We include the template here to make the probe-generation procedure reproducible, since probe quality affects the behavioral comparison used by PCR.

\begin{tcolorbox}[
    breakable,
    enhanced,
    colback=gray!3,
    colframe=black!35,
    boxrule=0.4pt,
    arc=1pt,
    left=4pt,
    right=4pt,
    top=4pt,
    bottom=4pt,
    title=Probe Input Generation Prompt Template,
    fonttitle=\bfseries\footnotesize,
    fontupper=\footnotesize,
    fontlower=\footnotesize,
    before skip=4pt,
    after skip=6pt
]
\noindent\textbf{System message.}\\
You are an expert at designing test inputs for competitive programming problems.
You generate test inputs that are strictly format-correct, satisfy all stated constraints, and are structurally distinct from any inputs already produced.

\tcblower
\noindent\textbf{User message.}\\
Generate input \#\emph{\{iter\_idx\}} of \emph{\{k\_total\}} for the given competitive-programming problem.

\smallskip
\noindent\textbf{Inputs provided to the prompt.}
\begin{itemize}
    \item \emph{\{problem\_content\}}: the full problem statement.
    \item \emph{\{prior\_inputs\_json\}}: all previously generated probe inputs for this problem.
\end{itemize}
The final answer must be a raw input string, not a JSON object.

\smallskip
\noindent\textbf{Generation requirements.}
\begin{itemize}
    \item Follow the input format specified by the problem statement, including quoting, newlines, list nesting, and any outer test-case count line.
    \item If the format uses an outer test-case count line, such as $t$, $T$, or $q$ on the first line, begin with a valid count and include exactly that number of case blocks.
    \item Make the new input structurally different from both the problem-statement examples and all prior generated inputs.
\end{itemize}

\smallskip
\noindent\textbf{Required output structure.}
\begin{enumerate}
    \item \textbf{Reasoning}: identify an uncovered aspect of the problem, such as size, range, sparsity, pattern, character set, distribution, or adversarial structure.
    \item \textbf{Format and constraint check}: state whether an outer test-case count line is used, state the generated count when applicable, and verify all size, value, structural, and format constraints.
    \item \textbf{Distinctness verification}: for each prior input, give a concrete comparison of the form ``Prior X had A, mine has B, and A differs from B on dimension Y.'' Use concrete values rather than abstract claims.
    \item \textbf{Final Input}: output the single raw input string in canonical format. If an outer test-case count line is used, the raw input must begin with that count line.
\end{enumerate}

\smallskip
\noindent\textbf{Size constraint.}
The raw input string must not exceed $5000$ characters.
For problems with very large limits, use representative sizes between $100$ and $1000$ rather than enumerating inputs at the maximum scale.
\end{tcolorbox}

\begin{table*}[t!]
  \centering
  \small
  \setlength{\tabcolsep}{4.0pt}
  \renewcommand{\arraystretch}{1.12}
  \caption{Transfer results for Qwen3-4B adapted on CodeContests (CC) and evaluated on LiveCodeBench (LCB), CodeForces (CF), and TACO. The CC columns report in-domain transductive adaptation; the remaining columns report zero-shot transfer from the final CC-adapted checkpoint. All values are percentages.}
  \label{tab:cc_transfer_4b}
  \begin{tabular}{@{}llcc@{\hspace{10pt}}*{8}{c}@{}}
    \toprule
    & & \multicolumn{2}{c}{In-Domain Adapt.} & \multicolumn{8}{c}{Zero-Shot Transfer} \\
    \cmidrule(lr){3-4}\cmidrule(l){5-12}
    Model & Method & \multicolumn{2}{c}{CC} & \multicolumn{2}{c}{LCB} & \multicolumn{2}{c}{CF} & \multicolumn{2}{c}{TACO} & \multicolumn{2}{c}{Mean} \\
    \cmidrule(lr){3-4}\cmidrule(lr){5-6}\cmidrule(lr){7-8}\cmidrule(lr){9-10}\cmidrule(l){11-12}
    & & P@1 & P@16 & P@1 & P@16 & P@1 & P@16 & P@1 & P@16 & P@1 & P@16 \\
    \midrule
    Qwen3-4B & Base                  & 25.13 & 42.26 & \underline{26.25} & 36.00 & 9.81 & 21.84 & 6.09 & 13.28 & 14.05 & 23.71 \\
    & GRPO$_{\mathrm{Pub}}$ & \underline{40.13} & 44.12 & 25.61 & 32.57 & 10.71 & 22.48 & 6.09 & 12.94 & \underline{14.14} & 22.66 \\
    & GRPO$_{\mathrm{PCR}}$ & 27.05 & 34.45 & 25.54 & 36.00 & 9.89 & 23.77 & 5.99 & 12.71 & 13.81 & 24.16 \\
    & NSR$_{\mathrm{PCR}}$  & 27.42 & \underline{50.42} & 24.61 & \underline{39.43} & \underline{10.95} & \textbf{32.33} & \underline{6.73} & \underline{17.03} & 14.10 & \underline{29.60} \\
    \rowcolor{oursrow}
    & ERPO                  & \textbf{41.62} & \textbf{57.56} & \textbf{33.75} & \textbf{45.14} & \textbf{15.81} & \underline{30.62} & \textbf{11.78} & \textbf{20.32} & \textbf{20.45} & \textbf{32.03} \\
    \bottomrule
  \end{tabular}
\end{table*}

\section{Transfer Results after CodeContests Adaptation (Qwen3-4B)}
\label{sec:cc_transfer_4b}

To complement the main transfer results (Table~\ref{tab:main_results}), which use LiveCodeBench as the adaptation source, we additionally adapt Qwen3-4B on the unlabeled CodeContests problem statements and transfer the resulting checkpoint to the other three code benchmarks, as reported in Table~\ref{tab:cc_transfer_4b}. The CodeContests columns report in-domain transductive adaptation, while the remaining columns report zero-shot transfer without target-benchmark adaptation. We include NSR$_{\mathrm{PCR}}$ because its checkpoint is also used in Table~\ref{tab:inference_time_scaling}. As in the main experiments, we evaluate the final checkpoint from the prespecified adaptation schedule without hidden-test-based checkpoint selection.

Table~\ref{tab:cc_transfer_4b} shows that the main conclusion is not specific to LiveCodeBench adaptation. ERPO performs best on both in-domain metrics and achieves the best mean transfer pass@1 and pass@16. GRPO$_{\mathrm{Pub}}$ improves in-domain pass@1 using public test cases, but this gain does not transfer, while direct PCR optimization provides no consistent benefit. NSR$_{\mathrm{PCR}}$ mainly improves pass@16 without comparable gains in pass@1. This pattern agrees with Table~\ref{tab:main_results}: ERPO is the only method that consistently strengthens both single-pass accuracy and the pass@k frontier.

\begin{figure*}[t!]
  \centering
  \includegraphics[width=\textwidth]{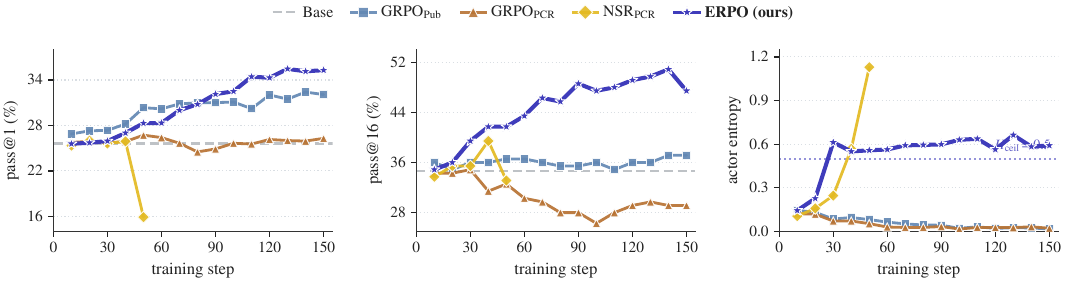}
  \caption{Per-step training dynamics on Qwen3-8B / LiveCodeBench. Left: pass@1. Middle: pass@16. Right: actor policy entropy. Dashed gray lines mark the Base reference. The dotted indigo line marks ERPO's fixed entropy ceiling $H_{\mathrm{ceil}}=0.5$. NSR$_{\mathrm{PCR}}$ is shown through step 50 because its entropy subsequently diverges and validation pass@k collapses.}
  \label{fig:training_trajectories}
\end{figure*}
\begin{table*}[t!]
  \centering
  \small
  \setlength{\tabcolsep}{3.0pt}
  \renewcommand{\arraystretch}{1.12}
  \caption{Inference-time pass@k on Qwen3-8B adapted on LiveCodeBench, evaluated on the LiveCodeBench test split. All values are percentages.}
  \label{tab:scaling_8b_lcb}
  \begin{tabular}{@{}l*{9}{C{0.072\textwidth}}@{}}
    \toprule
    Method & P@1 & P@2 & P@4 & P@8 & P@16 & P@32 & P@64 & P@128 & P@256 \\
    \midrule
    Base                  & 25.46 & 28.07 & 30.55 & 32.56 & 34.04 & 35.45 & 36.99 & 38.56 & 40.00 \\
    GRPO$_{\mathrm{Pub}}$ & \underline{30.47} & \underline{31.83} & \underline{32.96} & 34.09 & 35.35 & 36.66 & 37.92 & 39.25 & 40.57 \\
    GRPO$_{\mathrm{PCR}}$ & 26.13 & 27.54 & 28.80 & 29.94 & 31.00 & 32.07 & 33.20 & 34.61 & 36.57 \\
    NSR$_{\mathrm{PCR}}$  & 25.40 & 29.14 & 32.28 & \underline{35.11} & \underline{37.83} & \underline{40.44} & \underline{42.96} & \underline{45.49} & \underline{48.00} \\
    \rowcolor{oursrow}
    ERPO                  & \textbf{34.73} & \textbf{39.03} & \textbf{42.69} & \textbf{45.89} & \textbf{48.56} & \textbf{50.84} & \textbf{52.99} & \textbf{55.09} & \textbf{57.14} \\
    \bottomrule
  \end{tabular}
\end{table*}

\section{Training Trajectory Analysis}
\label{sec:training_trajectories}

Figure~\ref{fig:training_trajectories} shows the training dynamics behind the main results. GRPO$_{\mathrm{Pub}}$ improves pass@1, but its pass@16 remains close to the Base reference and its entropy declines rapidly, suggesting a narrower solution distribution without a stronger pass@k frontier. Direct PCR optimization does not improve pass@1 and steadily reduces pass@16, consistent with Table~\ref{tab:pcr_score_analysis}, where high PCR is not a reliable correctness label. NSR$_{\mathrm{PCR}}$ shows the opposite failure mode: it initially improves pass@16, but pass@1 declines as entropy grows and validation pass@k eventually collapses. ERPO avoids both patterns, sustaining improvements in pass@1 and pass@16 while keeping entropy stable. Its rank mask avoids positive updates on high-PCR candidates, while the retained low-PCR samples typically provide negative updates. The entropy ceiling further limits upward entropy drift. These dynamics are consistent with the joint gains in Table~\ref{tab:main_results}.

\section{Inference-Time Scaling after LiveCodeBench Adaptation (Qwen3-8B)}
\label{sec:scaling_8b_lcb}

To assess whether the inference-time scaling behavior in Table~\ref{tab:inference_time_scaling} extends to a larger model capacity, we report pass@k for Qwen3-8B adapted on LiveCodeBench, using up to $N=256$ samples per problem.

Table~\ref{tab:scaling_8b_lcb} shows that ERPO leads at every reported sampling budget and that its advantage over the Base model grows with $k$. Its pass@1 also exceeds the Base model's pass@16 with a single output. GRPO$_{\mathrm{Pub}}$ improves the small-$k$ region but saturates near the Base frontier at large $k$, while GRPO$_{\mathrm{PCR}}$ falls below the Base model beyond pass@1. NSR$_{\mathrm{PCR}}$ leaves pass@1 nearly unchanged while improving large-$k$ performance, suggesting that its gains mainly reflect sampling diversity. ERPO improves both single-pass accuracy and the high-$k$ frontier, showing that its rank-masked update and entropy ceiling remain effective at the larger model capacity.

\section{Results across Independent Runs}
\label{sec:independent_runs}

To better demonstrate the robustness of ERPO's performance gains, we repeated the complete Qwen3-4B setting reported in Table~\ref{tab:main_results} using three independent end-to-end runs for each adaptation method on LCB.
Each resulting checkpoint was evaluated in-domain on LCB and zero-shot on CC, CF, and TACO.
We report the mean and standard deviation across these runs in Table~\ref{tab:independent_runs}.
For the Base model, which does not involve training, the mean and standard deviation are computed across three independent evaluation runs.
Across the three runs, ERPO achieves the highest mean on all eight evaluation metrics.
Its margins over the strongest baselines are consistently larger than the corresponding standard deviations, indicating that the improvements are robust to the observed run-to-run variation.

\begin{table*}[t]
  \centering
  \footnotesize
  \setlength{\tabcolsep}{1.5pt}
  \renewcommand{\arraystretch}{1.12}
  \caption{Results across three independent runs in the Qwen3-4B setting adapted on LiveCodeBench. Each entry reports the mean $\pm$ standard deviation. Values are percentages.}
  \label{tab:independent_runs}
  \begin{tabular}{@{}l*{8}{c}@{}}
    \toprule
    & \multicolumn{2}{c}{LCB} & \multicolumn{2}{c}{CC} & \multicolumn{2}{c}{CF} & \multicolumn{2}{c}{TACO} \\
    \cmidrule(lr){2-3}\cmidrule(lr){4-5}\cmidrule(lr){6-7}\cmidrule(l){8-9}
    Method & P@1 & P@16 & P@1 & P@16 & P@1 & P@16 & P@1 & P@16 \\
    \midrule
    Base & $26.04\pm0.42$ & $34.43\pm0.87$ & $24.94\pm0.17$ & $41.14\pm0.97$ & $9.68\pm0.12$ & $22.48\pm1.11$ & $6.27\pm0.11$ & $13.24\pm0.24$ \\
    GRPO$_{\mathrm{Pub}}$ & $33.45\pm0.27$ & $35.61\pm0.45$ & $26.39\pm0.15$ & $40.03\pm0.87$ & $10.72\pm0.05$ & $23.48\pm0.81$ & $6.01\pm0.13$ & $12.98\pm0.63$ \\
    GRPO$_{\mathrm{PCR}}$ & $27.28\pm0.22$ & $29.54\pm0.62$ & $26.04\pm0.11$ & $36.54\pm0.24$ & $8.41\pm0.10$ & $20.20\pm0.12$ & $5.58\pm0.12$ & $10.03\pm0.36$ \\
    NSR$_{\mathrm{PCR}}$ & $24.90\pm0.32$ & $38.36\pm1.27$ & $27.97\pm0.57$ & $51.67\pm0.89$ & $10.60\pm0.17$ & $29.76\pm0.61$ & $6.86\pm0.13$ & $16.69\pm0.80$ \\
    \rowcolor{oursrow}
    ERPO & $36.64\pm0.33$ & $46.96\pm1.15$ & $42.13\pm0.47$ & $58.44\pm0.24$ & $20.64\pm0.29$ & $38.26\pm0.33$ & $13.42\pm0.09$ & $21.30\pm0.40$ \\
    \bottomrule
  \end{tabular}
\end{table*}

\section{Results on More Benchmarks}
\label{sec:more_benchmark_results}

We conducted additional in-domain test-time adaptation experiments on BigCodeBench (BCB; \citealp{zhuo2025bigcodebench}) and DS-1000 \citep{lai2023ds1000}.
Each method was trained on the unlabeled problems of the corresponding benchmark using the same Qwen3-4B backbone, adaptation budget, and optimization settings as in the main experiments.
Table~\ref{tab:more_benchmark_results} reports the results.

\begin{table}[t]
  \centering
  \small
  \setlength{\tabcolsep}{4.0pt}
  \renewcommand{\arraystretch}{1.12}
  \caption{In-domain adaptation results on two additional benchmarks using Qwen3-4B. Parentheses show absolute changes from Base. Values are percentages.}
  \label{tab:more_benchmark_results}
  \begin{tabular}{@{}llcc@{}}
    \toprule
    Benchmark & Method & P@1 & P@16 \\
    \midrule
    BCB & Base & 35.5 & 45.4 \\
    & GRPO$_{\mathrm{PCR}}$ & 36.7 $(+1.2)$ & 44.3 $(-1.1)$ \\
    & NSR$_{\mathrm{PCR}}$ & 30.4 $(-5.1)$ & 49.4 $(+4.0)$ \\
    \rowcolor{oursrow}
    & ERPO & \textbf{39.2 $(+3.7)$} & \textbf{57.8 $(+12.4)$} \\
    \midrule
    DS-1000 & Base & 22.7 & 29.2 \\
    & GRPO$_{\mathrm{PCR}}$ & 23.8 $(+1.1)$ & 28.9 $(-0.3)$ \\
    & NSR$_{\mathrm{PCR}}$ & 17.2 $(-5.5)$ & 36.3 $(+7.1)$ \\
    \rowcolor{oursrow}
    & ERPO & \textbf{27.4 $(+4.7)$} & \textbf{44.9 $(+15.7)$} \\
    \bottomrule
  \end{tabular}
\end{table}

ERPO consistently improves both pass@1 and pass@16 on both benchmarks.

\section{Sensitivity to Rank-Mask Percentile}
\label{sec:rank_mask_sensitivity}

We selected 50\% as a simple median split because it avoids an absolute PCR threshold and task-specific tuning without labeled validation data.
To assess sensitivity to this choice, we vary the masked percentile on LiveCodeBench while keeping all other experimental settings fixed.
Table~\ref{tab:rank_mask_sensitivity} reports P@1 and P@16 for both backbones.

\begin{table}[H]
  \centering
  \small
  \setlength{\tabcolsep}{2.5pt}
  \renewcommand{\arraystretch}{1.12}
  \caption{Sensitivity to the rank-mask percentile on LiveCodeBench. Values are percentages.}
  \label{tab:rank_mask_sensitivity}
  \begin{tabular}{@{}llccccc@{}}
    \toprule
    Backbone & Metric & Base & 10\% & 25\% & 50\% & 75\% \\
    \midrule
    Qwen3-4B & P@1  & 26.0 & 29.8 & 36.9 & 36.7 & 36.4 \\
              & P@16 & 34.4 & 36.6 & 46.3 & 46.3 & 46.3 \\
    Qwen3-8B & P@1  & 25.6 & 30.0 & 35.8 & 35.1 & 34.9 \\
              & P@16 & 34.6 & 37.7 & 47.4 & 50.9 & 49.7 \\
    \bottomrule
  \end{tabular}
\end{table}

Performance remains relatively stable across masking ratios from 25\% to 75\%, and all three settings substantially improve over the corresponding Base models.
Masking only 10\% still improves over the Base models but performs worse than masking 25\%--75\%, likely because more candidates with unreliable positive advantages remain unmasked.

\section{Sensitivity to Generated Probe Count}
\label{sec:probe_count_sensitivity}

Let $N_{\mathrm{gen}}$ denote the number of additional probe inputs generated per problem.
We use $N_{\mathrm{gen}}=10$ by default to balance PCR reliability and probe-execution cost.
To assess sensitivity to this choice, we first evaluate how the number of generated probes affects PCR AUC for predicting whether a candidate passes the hidden tests on LiveCodeBench with Qwen3-4B.

\begin{table}[H]
  \centering
  \small
  \setlength{\tabcolsep}{5.0pt}
  \renewcommand{\arraystretch}{1.12}
  \caption{PCR AUC for different numbers of generated probes on LiveCodeBench with Qwen3-4B.}
  \label{tab:probe_count_auc}
  \begin{tabular}{@{}lccccc@{}}
    \toprule
    $N_{\mathrm{gen}}$ & 1 & 3 & 5 & 10 & 20 \\
    \midrule
    PCR AUC & 0.706 & 0.754 & 0.774 & 0.790 & 0.799 \\
    \bottomrule
  \end{tabular}
\end{table}

PCR AUC improves as $N_{\mathrm{gen}}$ increases, indicating that additional probes produce a more reliable consensus signal.
The improvement diminishes beyond $N_{\mathrm{gen}}=10$, with AUC increasing by only 0.009 when the generated probe count is doubled from 10 to 20.
We further evaluate the effect of $N_{\mathrm{gen}}$ on downstream performance in Table~\ref{tab:probe_count_performance}.

\begin{table}[H]
  \centering
  \small
  \setlength{\tabcolsep}{3.0pt}
  \renewcommand{\arraystretch}{1.12}
  \caption{Sensitivity to the generated probe count on LiveCodeBench. Values are percentages.}
  \label{tab:probe_count_performance}
  \begin{tabular}{@{}llcccc@{}}
    \toprule
    Backbone & Metric & Base & \multicolumn{3}{c}{$N_{\mathrm{gen}}$} \\
    \cmidrule(l){4-6}
             &        &      & 5 & 10 & 20 \\
    \midrule
    Qwen3-4B & P@1  & 26.0 & 35.8 & 36.7 & 36.5 \\
              & P@16 & 34.4 & 45.7 & 46.3 & 46.9 \\
    Qwen3-8B & P@1  & 25.6 & 34.5 & 35.1 & 35.5 \\
              & P@16 & 34.6 & 48.0 & 50.9 & 50.3 \\
    \bottomrule
  \end{tabular}
\end{table}

Performance remains stable across $N_{\mathrm{gen}}\in\{5,10,20\}$, and increasing $N_{\mathrm{gen}}$ from 10 to 20 does not yield consistent improvements despite the additional execution cost.
These results support $N_{\mathrm{gen}}=10$ as a practical balance between reward reliability, downstream performance, and computational cost.

\section{Probe Examples and Hidden-Test Correctness}
\label{sec:probe_examples}

We provide two representative cases illustrating PCR's asymmetric reliability.

\paragraph{Case A: Probe disagreement reveals incorrect candidates.}
Consider a task that counts arrays whose elements satisfy given bounds while preserving the adjacent differences of a reference array.
One generated probe is:
\begin{tcolorbox}[
    colback=gray!3,
    colframe=black!35,
    boxrule=0.4pt,
    arc=1pt,
    left=3pt,
    right=3pt,
    top=3pt,
    bottom=3pt
]
\small\ttfamily
original = [1, 3, 2, 4, 3]\\
bounds = [[1, 3], [2, 4], [1, 3],\\
\phantom{bounds = [}[3, 5], [2, 4]]
\end{tcolorbox}

The outputs of the 16 candidates form four clusters:
\begin{itemize}
    \setlength{\itemsep}{0pt}
    \setlength{\parsep}{0pt}
    \setlength{\topsep}{2pt}
    \item \texttt{2}: 7 candidates, all passing the hidden tests;
    \item \texttt{3}: 4 candidates, all failing;
    \item \texttt{0}: 3 candidates, all failing;
    \item \texttt{1}: 2 candidates, all failing.
\end{itemize}
The feasible first-element values are 1 and 2, so the correct answer is 2.
Across the complete set of ten generated probes, all seven candidates that pass the hidden tests receive PCR $=1.0$, whereas all nine failing candidates receive lower PCR.
Thus, PCR perfectly separates passing and failing candidates in this case.

\paragraph{Case B: Unanimous probe agreement misses a boundary error.}
Consider a task that computes $X=1+N+N^{2}+\cdots+N^{M}$ and outputs \texttt{inf} when $X>10^{9}$.
All 16 candidates agree on all ten generated probes, including \texttt{2 30} $\rightarrow$ \texttt{inf} and \texttt{2 20} $\rightarrow$ \texttt{2097151}, and therefore receive PCR $=1.0$.
However, two candidates fail the hidden input \texttt{2 29}, where $X=2^{30}-1=1073741823>10^{9}$.
They output \texttt{1073741823} instead of \texttt{inf}.
The probes miss this narrow boundary region, so the incorrect candidates remain indistinguishable from the correct ones.

Together, these cases illustrate why ERPO uses PCR asymmetrically: low consensus can provide useful negative evidence, whereas high consensus is not treated as positive supervision.

\end{document}